\documentclass[a4paper,fleqn]{cas-sc}
\usepackage{apacite}
\usepackage{multirow}
\usepackage{subfigure}
\usepackage{color}
\usepackage{subcaption}
\usepackage[numbers]{natbib}
\usepackage{bbding}
\usepackage{amsmath}
\usepackage{booktabs, tabularx}
\usepackage{caption}
\usepackage[numbers]{natbib}
\usepackage{etoolbox}
\usepackage{natbib}
\usepackage{hyperref,url}
\newcounter{bibcount}
\makeatletter
\patchcmd{\@lbibitem}{\item[}{\item[\hfil\stepcounter{bibcount}{\thebibcount.}}{}{}
\renewcommand\NAT@bibsetup%
   [1]{\setlength{\leftmargin}{\bibhang}\setlength{\itemindent}{-\parindent}%
       \setlength{\itemsep}{\bibsep}\setlength{\parsep}{\z@}}
\makeatother
\begin{document}

% \begin{titlepage}
    
%         \begin{center}
%         \vspace{16\baselineskip}
%         {\Large Benchmarking Foundational Models on the Arabic MentalQA Dataset\par}

%         \vspace{8\baselineskip}
%         {{Hassan Alhuzali$^1$\\Ashwag Alasmari$^2$}}
%         \end{center}
        
%         \vspace{8\baselineskip}
%         \noindent $^1$Department of Computer Science and Artificial Intelligence, Umm Al-Qura University, Makkah, Saudi Arabia.\\
%         $^2$Department of Computer Science, King Khalid University, Abha, Saudi Arabia.

%         \vspace{2\baselineskip}
%         \noindent\textbf{*Corresponding author}: Hassan Alhuzali
%         \\
%         Email Address: \\
%         hrhuzali@uqu.edu.sa
%         \\
%         aasmry@kku.edu.sa

%     \end{titlepage}

\ExplSyntaxOn
\cs_gset:Npn \__first_footerline:
  { \group_begin: \small \sffamily \__short_authors: \group_end: }
\ExplSyntaxOff

\let\WriteBookmarks\relax
\def\floatpagepagefraction{1}
\def\textpagefraction{.001}

\shorttitle{}

\shortauthors{Alhuzali and Alasmari}

\title [mode = title]{Evaluating the Effectiveness of the Foundational Models for Q\&A Classification  in Mental Health care}                      



\author[1]{Hassan Alhuzali*}[orcid=0000-0002-0935-0774]
% \cormark[1]



\cortext[cor1]{Corresponding author}
\ead{halhuzali@uqu.edu.sa}
\affiliation[1]{organization={Department of Computer Science and Artificial Intelligence, Umm Al-Qura University},
    city={Makkah},
    country={Saudi Arabia}}
    
\credit{Methodology, Software, Validation, Formal analysis, Visualization, Writing - original draft}

\author[2]{Ashwag Alasmari}[orcid=]

\credit{Methodology and Writing - review \& editing}

% \author[2]{Hassan Alhuzali}[orcid=0000-0002-0935-0774]

% \affiliation[2]{organization={College of Computers and Information Systems, Umm Al-Qura University}, city={Makkah}, country={Saudi Arabia}}

% \credit{Visualization, Writing - review \& editing}

% \author[1]{Boyang Liu}[orcid=0000-0002-0814-002X]

% \credit{Writing - review \& editing}

% \author[1,3]{Sophia Ananiadou}[orcid=0000-0002-4097-9191]

\affiliation[2]{organization={Department of Computer Science, King Khalid University},
    city={Abha},
    country={Saudi Arabia}}
    
% \credit{Funding acquisition, Supervision, Writing - Review \& Editing}




\begin{abstract}
% Pre-trained Language Models (PLMs) have the potential to revolutionize mental health support by providing accessible and culturally sensitive resources. However, their effectiveness for Arabic language remains largely unexplored. This study addresses this gap by benchmarking MentalQA, the Arabic mental health question-answering dataset featuring conversational-style interactions. We experimented with three types of learning, i.e., using traditional feature extraction, PLMs as feature extractors and Fine-tuning PLMs. Arabic PLMs include AraBERT, CAMelBERT, and MARBERT. While traditional feature extractors with SVM achieved promising performance, PLMs demonstrated even better results due to their ability to capture semantic meaning. For instance, MARBERT achieved the best results with a Jaccard Score of 0.80 for question classification and a jaccard score of 0.86 for answer classification. To gain a deeper understanding of the experiments and factors influencing model performance, we conducted in-depth analysis including effects of fine-tuning versus non-fine-tuning, impact of varying data size, and error analysis. These findings suggest PLMs models hold promise for mental health support in Arabic, especially for facilitating intervention assistance and providing guidance for non-clinical applications.
Background: Pre-trained Language Models (PLMs) have the potential to transform mental health support by providing accessible and culturally sensitive resources. However, despite this potential, their effectiveness in mental health care and specifically for the Arabic language has not been extensively explored. To bridge this gap, this study evaluates the effectiveness of foundational models for classification of Questions and Answers (Q\&A) in the domain of mental health care. 

\noindent Methods: We leverage the MentalQA dataset, an Arabic collection featuring question-answering interactions related to mental health. In this study, we conducted experiments using four different types of learning approaches: traditional feature extraction, PLMs as feature extractors, Fine-tuning PLMs and prompting large language models (GPT-3.5 and GPT-4) in zero-shot and few-shot learning settings. We utilized Arabic PLMs such as AraBERT, CAMelBERT, and MARBERT. 

\noindent Results: traditional feature extractors combined with Support Vector Machines (SVM) showed promising performance, PLMs exhibited even better results due to their ability to capture semantic meaning. For example, MARBERT achieved the highest performance with a Jaccard Score of 0.80 for question classification and a Jaccard Score of 0.86 for answer classification. To attain a better understanding of the experiments and factors influencing model performance, we conducted an in-depth analysis including examining the effects of fine-tuning versus non-fine-tuning, the impact of varying data size, and conducting error analysis. Our analysis demonstrates that fine-tuning proved to be beneficial for enhancing the performance of PLMs, and the size of the training data played a crucial role in achieving high performance. We also explored prompting, where few-shot learning with GPT-3.5 yielded promising results. There was an improvement of 12\% for question 
classification and 45\% for answer classification. 

\noindent Conclusion: Based on our findings, it can be concluded that PLMs and prompt-based approaches hold promise for mental health support in Arabic, offering valuable resources for individuals seeking assistance in this domain\footnote{Upon acceptance, the source code will be made available on GitHub.}.

%Mental disorders detection on social media aims at early identification of tendency towards mental health problems from social media posts, which provides a potential way for social intervention.

\end{abstract}

% \includegraphics{figs/grabs.pdf}

% \begin{highlights}

% \item We conducted the first set of experiments on the MentalQA dataset, the novel Arabic mental health question-answering dataset featuring conversational-style interactions.

% \item We benchmarked classical machine learning models and Pre-trained Language Models (PLMs) on the MentalQA dataset.

% \item We demonstrated the current capabilities and limitations of both classical models and Arabic PLMs for mental health support and, ways for further improving the results of PLMs on the MentalQA dataset. By identifying the areas where these models excel and where they fall short, we can pave the way for further development and refinement.
    
% % \item Our experiments and analyses, we have identified the present capabilities and limitations of Arabic PLMs in supporting mental health, as well as identifying ways to further enhance PLMs performance on the Arabic MentalQA dataset.

% % \item ........
% \end{highlights}

% Keywords
% Each keyword is seperated by \sep Natural Language Processing, Foundation models, Mental Health, Multi-label Classification, Q/A Classification.

\begin{keywords}
Mental health \sep Natural language processing \sep  Question/answer classification \sep Text classification \sep Large Language Models
\end{keywords}

\maketitle

\section{Introduction}\label{sec:introduction}

Mental health disorders are a significant global burden, affecting nearly one billion people across all demographics according to the World Health Organization \cite{world2022world}.  These conditions account for a substantial 13\% of the global disease burden.  Despite this widespread prevalence, access to effective care remains limited.  Only half of those with mental health disorders receive treatment, and even fewer receive adequate care \cite{who2004prevalence}.  The economic impact is also significant, with depression and anxiety alone costing the global economy around \$1 trillion annually \cite{marquez2016making}. Natural Language Processing (NLP) offers promising solutions for addressing these challenges, particularly in early intervention and resource allocation \cite{le2021machine}. Recent PLMs have shown exceptional efficacy in various NLP tasks, including classification, language synthesis, and question answering \cite{devlin2019bert}. This paves the way for the development of innovative mental health support systems that leverage the capabilities of NLP.

PLMs have made significant advances in recent years, revolutionizing applications across different domains including medicine \cite{he2023survey}.  However, research dedicated to understanding and enhancing PLMs for mental health  remains in its early stages. The application of PLMs in mental health holds exciting possibilities for both patients seeking support and health care providers aiming to enhance their services. Patient applications offer a spectrum of experiences: immersive conversation directly with the model such as \cite{liu2023chatcounselor, brocki2023deep}, or using the PLMs to understand and categorize user input for connection with a human therapist such as \cite{sharma2023human}. For health care providers, PLMs can generate recommendations or suggested answers to utilize. 

However, the potential effectiveness of PLMs in the mental health domain depends on their ability to understand the nuances of human language including the subjective, variable nature of mental health symptoms and the need for specialized skills. This becomes even more challenging when considering languages like Arabic. With over 400 million native speakers worldwide, Arabic language stands as a language of great richness and complexity \cite{guellil2021arabic}. Arabic is among the top five spoken languages in the world. In addition, with over 280 million users worldwide, Arabic language ranks as the fourth most used language on the internet~\cite{guellil2021arabic}.

Despite efforts worldwide in other languages~\cite{atapattu2022emoment, kabir2022detection, sun-etal-2021-psyqa}, the Arabic language is an understudied language regarding mental health disorders. According to~\cite{zhang2022natural}, there is a significant disparity in the availability of mental health datasets across different languages. The study reveals that English datasets dominate, accounting for 81\% of the total, while Chinese datasets follow at 10\%. In contrast, Arabic datasets are notably scarce, representing only 1.5\% of the dataset resources. To date, only a handful of works have considered mental health issues in the Arabic language~\cite{ abdulsalam2024detecting,aldhafer2022depression, al2022depression, al2021monitoring, 9694178}. In particular,~\cite{aldhafer2022depression} developed depression detection models from Arabic texts on Twitter which focused on the cultural stigma surrounding depression in Arab societies. Another study by~\cite{al2022depression} also focused on the detection of depression in which they applied various machine learning algorithms and feature extraction techniques. 

\subsection{Research Objectives and Contributions}

The primary goal of this study is to perform comprehensive experiments on the recently developed dataset called Arabic MentalQA~\cite{alhuzali2024mentalqa}. In light of this objective, we present the following contributions: \begin{itemize}
    \item Conducting the first set of experiments on the MentalQA dataset, the novel Arabic mental health question-answering dataset featuring question-answering interactions.
    
    \item Exploring the effectiveness of  classical machine learning models and PLMs  on the MentalQA dataset.

    \item Demonstrating the current capabilities and limitations of both classical models, Arabic PLMs and promoting large language models for mental health care and, ways for further improving the results of PLMs on the MentalQA dataset. By identifying the areas where these models excel and where they fall short, we can pave the way for further development and refinement.
\end{itemize}

This research will contribute valuable insights into the feasibility of utilizing PLMs for Arabic mental health support, ultimately aiding in the creation of accessible and culturally sensitive resources for Arabic-speaking populations. The rest of this manuscript is organized as follows: Section~\ref{relatedwork} reviews relevant works in the field. Section~\ref{exper} details the conducted experiments and the implementation details. Section~\ref{results} presents the results of our experiments, while section~\ref{analysis} focuses on the evaluation of results, examining factors influencing model performance. Section~\ref{disc} discuses the results and implications, as well as highlighting some limitations and ethical considerations. Finally, Section~\ref{conc} summarizes our work, proposes potential avenues for future research. 

%3) Demonstrating the current capabilities and limitations of Arabic PLMs in supporting mental health and ways for further improving the results of PLMs on the Arabic MentalQA dataset.

% Through our results and analyses, we have identified the present capabilities and limitations of Arabic PLMs in supporting mental health, as well as showing ways to further enhance PLMs performance on the Arabic MentalQA dataset.

% \begin{table}[]
% \caption{Question (Q) and Answer (A) types.}
% \label{tab:my-table}
% \begin{tabular}{llll}
% \toprule
% \# & Q-Types                & \# & A-Types           \\ \midrule
% 1 & Diagnosis              & 1  & Information       \\
% 2  & Treatment              & 2  & Direct Guidance   \\
% 3  & Anatomy                & 3  & Emotional Support \\
% 4  & Epidemiology           &    &                   \\
% 5  & Healthy Lifestyle      &    &                   \\
% 6  & Health Provider Choice &    &                   \\
% 7  & Other                  &    & \\   
% \bottomrule
% \end{tabular}
% \end{table}

\section{Related Work}\label{relatedwork}
Significant research efforts have been devoted to utilizing computational methods for mental health applications. This section delves into the relevant literature across three key domains. Firstly, we examine the established approaches of Classical Machine Learning (ML) techniques used for mental health tasks. Secondly, we explore the emerging field of PLMs and their potential for mental health applications. Finally, we analyze existing Resources for Mental Health Support.

\subsection{Classical Machine Learning}
A substantial body of research has explored computational methods for mental health applications.  Classical Machine Learning (ML) techniques have been a fundamental of this endeavor, offering a diverse set of algorithms for various tasks. Several classical ML algorithms have demonstrated promising results in mental health applications, each with its own strengths and weaknesses. Techniques like Support Vector Machines (SVMs) excel at classifying text data for tasks like depression detection, but lack interpretability \cite{fernandez2014we}. Decision Trees offer a clearer view of decision-making but can become unstable with complex data \cite{nikam2015comparative}. Naïve Bayes is efficient but has limitations with intricate datasets. Despite their success, classical ML often requires extensive feature engineering, hindering their ability to handle unstructured text data. This paves the way for PLMs – the focus of the next section – which can automatically learn these features from vast amounts of data.

\subsection{Pre-trained Language Models}

Significant progress has been made in developing PLMs for the English language, demonstrating their effectiveness in various Natural Language Processing (NLP) tasks. These powerful PLMs are trained on massive volumes of textual data, allowing them to perform a wide range of tasks. Prior models like ULMFiT~\cite{howard2018universal} excel in text classification tasks across languages, while BERT~\cite{devlin2019bert} established a foundation for ``deep bidirectional pre-training'', achieving SOTA performance on various sentence-level tasks. Subsequent advancements addressed limitations of BERT. ALBERT tackles memory and training time issues with parameter reduction techniques~\cite{lan2020albert} , while RoBERTa refines the pre-training process for improved performance~\cite{liu2019roberta}. XLNet goes further, addressing shortcomings in BERT's handling of masked positions and pre-training approaches~\cite{yang2019xlnet}. These advancements showcase the continuous development and growing capabilities of general domain PLMs for English NLP tasks.

Despite being the fourth most prevalent language online with more than 400 million speakers across 22 countries \cite{guellil2021arabic}, only a few attempts have been made to construct PLMs in Arabic. This gap can be attributed to the inherent complexity of Arabic, characterized by its rich morphology, diverse dialects, and unique writing system that reads from right to left, lacks capitalization, and utilizes character shapes that vary based on their position within a word \cite{shaalan2019challenges}.

One of the first efforts to construct a pre-trained LLM specifically for Arabic was hULMonA \cite{eljundi2019hulmona}, which is the first universal language model specifically designed for Arabic. This model obtained SOTA performance in Arabic sentiment analysis tasks, demonstrating its effectiveness on various Arabic datasets. AraBERT, another early effort, pre-trained the BERT model for Arabic, achieving advanced results in sentiment analysis, entity recognition, and question answering~\cite{antoun2020arabert}. Subsequent studies explored multilingual models like GigaBERTs and focused on the impact of data sources like OSCAR and Gigaword~\cite{lan2020empirical}. ARBERT and MARBERT are two models that were built by~\cite{abdul2020arbert}, and they were trained on a large number of Modern Standard and Dialect Arabic datasets, respectively. Another model by~\cite{inoue2021interplay} pre-train a single BERT-base model called CAMeLBERT, based on variations of dialect and classic Arabic data.~\cite{almazrouei-etal-2023-alghafa} introduced ``AlGhafa'', which is focused on benchmarking Arabic PLMs for multiple-choice evaluation. An additional work by~\cite{abdelali-etal-2024-larabench} conducts benchmarking recent advancements in PLMS against state-of-the-art models on various NLP tasks.  

\subsection{Resources for Mental Health Support}

While PLMs have been primarily focused on general domains, there's a growing interest in creating models tailored for specific domains like health. Researchers have developed several domain-specific PLMs for learning text representations in the health domain, such as BioMedical BERT~\cite{lee2020biobert} and Clinical BERT~\cite{alsentzer2019publicly}. These models excel in capturing domain-specific nuances compared to general-purpose PLMs. However, PLMs designed specifically for mental health remain scarce.  Pioneering efforts like MentalBERT and MentalRoBERTa address this gap, offering valuable resources for the mental health care research community~\cite{ji2022mentalbert}.~\cite{guo2024large} conducts a comprehensive review of existing work, highlighting the potential benefits and limitations of PLMs in early screening, digital interventions, and other clinical applications in mental health.

The efficacy of PLMs is demonstrably dependent upon the quality and quantity of data employed for their training. In literature, efforts were made to develop benchmark corpora for the mental health support \cite{coppersmith2015clpsych, alasmari2023chq, shen2017depression, turcan2019dreaddit, rastogi2022stress, boonyarat2024leveraging}. Existing examples include datasets for depression \cite{shen2017depression}, self-harm \cite{turcan2019dreaddit}, and stress \cite{rastogi2022stress}.  However, the field is evolving towards more nuanced datasets that capture emotions related to specific mental conditions. The CEASE dataset focuses on emotions of suicide attempters~\cite{ghosh-etal-2020-cease}, while EmoMent targets emotions linked to depression and anxiety~\cite{atapattu2022emoment}. Additionally, datasets are being developed for tasks like pain level identification in mental health notes \cite{chaturvedi2023identifying} and causal interpretation using datasets like CAMS \cite{garg2022cams}. Moreover, a recent study focused on wellness classification and extraction from Reddit posts related to mental health~\cite{garg2024wellxplain}. To accomplish this, the authors began by constructing a new corpus specifically tailored for this task. Subsequently, they evaluated the effectiveness of PLMs in analyzing the corpus. These advancements highlight a shift towards richer and more comprehensive mental health data resources, crucial for training effective PLMs in this domain.

%Current mental health datasets often target specific disorders like depression or anxiety \cite{coppersmith2015clpsych, shen2017depression, ghosh-etal-2020-cease, atapattu2022emoment, chaturvedi2023identifying, garg2022cams}. This narrow focus may limit AI models' ability to diagnose a broader range of mental health issues.

\subsection{How this Research Differs from Existing Work?}
While progress has been made, a significant gap exists in applying PLMs to Arabic-language mental health applications. This gap stems from two main challenges. The first challenge revolves around the scarcity of Arabic-language datasets, particularly those that are specifically tailored to the medical domain. The availability of such datasets is crucial for training PLMs to effectively comprehend and generate accurate content in the field of mental health. The second challenge lies in the limited availability of labeled data for mental health in the Arabic language. While the introduction of MentalQA, which is the first Arabic mental health dataset featuring question-answering interactions, has provided a valuable foundation, it is clear that more diverse and comprehensive datasets are required. Such datasets would enable the development of robust Arabic PLMs, capable of addressing a wider range of mental health applications. Given these limitations, we have been motivated to take action. Our objective is to benchmark the MentalQA dataset and utilize it as a stepping stone in supporting the development of powerful Arabic PLMs for mental health applications. Through this endeavor, we aim to bridge the existing gap and facilitate advancements in the field, ultimately benefiting individuals seeking mental health support in the Arabic-speaking world.

% First, there's a scarcity of Arabic-language datasets, particularly those specific to the medical domain. Second, the availability of labeled data for mental health is limited. While MentalQA, the first Arabic mental health dataset with conversational-style interactions, offers a valuable starting point, more diverse and comprehensive datasets are needed. These limitations motivated us to benchmark the MentalQA dataset to support the development of robust Arabic PLMs for mental health applications.

% In the realm of Arabic-language mental health applications, progress has undoubtedly been achieved. However, an evident gap persists when it comes to the utilization of Pre-trained Language Models (PLMs) in this context. This gap can be attributed to two primary challenges that need to be addressed.

\section{Experiments}\label{exper}

\subsection{Dataset Description}
\begin{figure}[h]
\centering
{\includegraphics[width=0.9\linewidth]{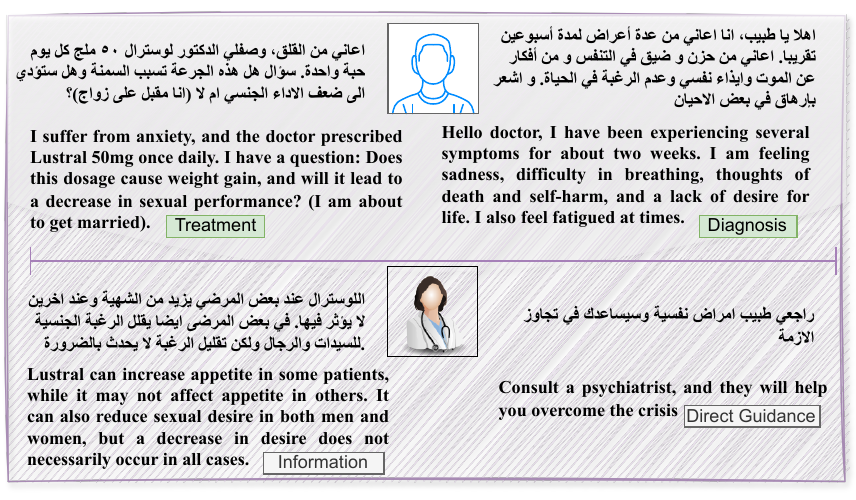}}
\caption{Example of two annotated Q\&A posts in MentalQA dataset, with each Q\&A post translated into English for better readability. The first row represents the questions, while the second row represents the corresponding answers. Additionally, the categories for each question and answer are included.}\label{ann_examples}
\end{figure}

In this work, we selected the MentalQA dataset~\cite{alhuzali2024mentalqa}, which consists of posts written in Arabic. This dataset was compiled from an Arabic medical platform called Altibbi, focusing on mental health questions and answers posted between 2020 and 2021. The MentalQA dataset comprises of 500 questions posted by patients, followed by 500 corresponding answers provided by professional doctors. Questions are classified into seven types: diagnosis, treatment, anatomy, epidemiology, healthy lifestyle, provider choice and other. Additionally, answers are classified into three strategies: information, direct guidance, and emotional support. Table~\ref{categories} further shows the complete range of categories that encompass both question and answer types within the dataset. To better illustrate the dynamics between patients and professional doctors, Figure~\ref{ann_examples} presents two annotated examples featuring different question and answer types. These examples serve as a visual representation of the interaction within the dataset and highlight the variety of exchanges that take place.

\subsection{Task Setting} 
The MentalQA dataset encompasses two tasks: the classification of question types and answer types. Both tasks allow for the assignment of multiple labels, employing a multi-label classification approach. To ensure proper evaluation, we divided the dataset into three subsets: a training set, accounting for 60\% of the total data, a validation set, representing 20\% of the total data, and a test set, also encompassing 20\% of the total data. To gain a thorough understanding of the MentalQA dataset, Table~\ref{data stats} provides a comprehensive overview of the dataset. This table includes information about the number of instances in the training, validation, and test sets, as well as the different types of questions and answers available in this dataset. 

\begin{table}[h]

\parbox{.45\linewidth}{
\caption{Question (Q) and Answer (A) types.}
\label{categories}
\scalebox{0.9}{
\begin{tabular}{llll}
\toprule
\# & Q-Types                & \# & A-Types           \\ \midrule
1 & Diagnosis              & 1  & Information       \\
2  & Treatment              & 2  & Direct Guidance   \\
3  & Anatomy                & 3  & Emotional Support \\
4  & Epidemiology           &    &                   \\
5  & Healthy Lifestyle      &    &                   \\
6  & Health Provider Choice &    &                   \\
7  & Other                  &    & \\   
\bottomrule
\end{tabular}}}
\hfill
\parbox{.45\linewidth}{
% \centering
\caption{Data statistics. }
\label{data stats}%
\scalebox{0.9}{
\begin{tabular}{lcc}
\toprule
\textbf{Info./Task} & \textbf{Q-Types} & \textbf{A-Types} \\ \midrule
{Train (\#)} & 300 & 300 \\
{Validation (\#)}  & 100 & 100 \\
{Test (\#)} & 100 & 100 \\
{Total (\#)} & 500 & 500 \\
{Classes} (\#)  & 7 & 3 \\
{Setup}  & multi-label & multi-label \\
\bottomrule
\end{tabular}}}
\end{table}%

% For a comprehensive overview of the MentalQA dataset, including the number of instances in the training, development, and test sets, as well as the number of question and answer types, please refer to Table~\ref{data stats}. Table~\ref{categories} presents the complete set of categories belonging to both question and answer types. Figure~\ref{ann_examples} presents two annotated examples with Q/A types. This illustrates the interaction between patient and professional doctors. 
% encompass various categories, namely:~\textit{(diagnosis, treatment, anatomy, epidemiology, health lifestyle, health provider choice, other}), whereas the answer types include the three following categories:~\textit{(information, direct guidance, and emotional support)}.

\subsection{Experimental Design}

\begin{figure}[h]
\centering
{\includegraphics[width=\linewidth]{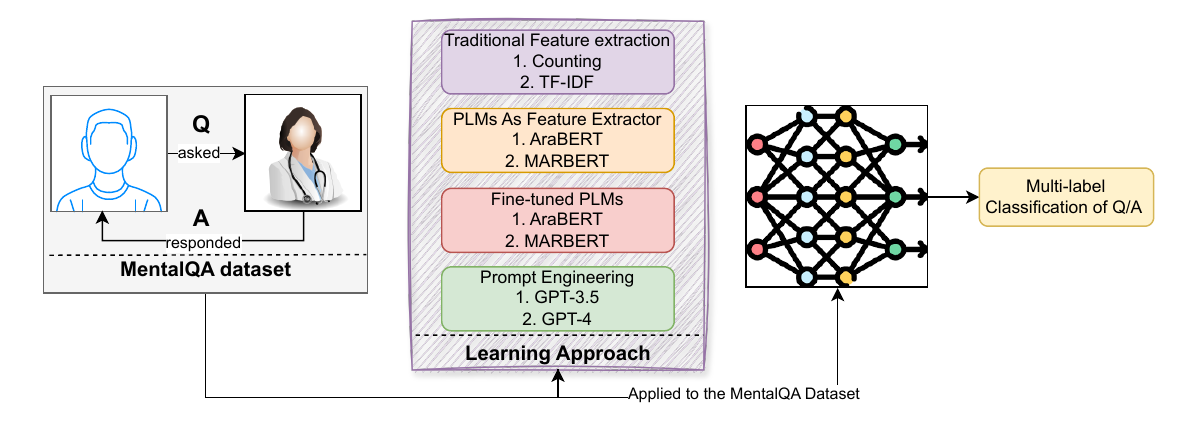}}
\caption{An overview of our experimental design. Specifically, it outlines the process by which the input is conveyed to the design of the learning approach, wherein the resulting outputs of various approaches are linked to the desired task outcome, namely, a multi-label classification of Q/A types.}\label{framework}
\end{figure}

In this paper, we explore four different approaches for the task of Q/A multi-label classification: traditional feature extraction methods with Support Vector Machines (SVM), utilizing PLMs as feature extractors with SVM, fine-tuning PLMs, and prompting engineering. Our goal is to assess the performance of these approaches specifically for the task of MentalQA dataset described in the above section. In addition, we include two common baselines that are based on the most frequent class and randomness.  Figure~\ref{framework} provides an illustration of our experimental design.

Traditional feature extraction methods have been widely used in text classification~\cite{peters2018deep,peters2018dissecting,peters2017semi}. These methods involve converting input text into numerical representations based on word or concept frequency or presence. Techniques like ``term frequency-inverse document frequency~\textit{(TF-IDF)}'' and counting methods have been extensively employed. These approaches offer simplicity and interpretability, relying on straightforward calculations. Subsequently, SVM, a popular machine learning algorithm, is employed to classify the transformed text data. The goal of SVM is to seek out an optimal hyperplane within the feature space, effectively segregating distinct classes. The effectiveness of this approach has been demonstrated in various studies~\cite{wahba2022comparison}.

The emergence of PLMs has revolutionized NLP tasks, including text classification~\cite{devlin2019bert, liu2019roberta,lan2020albert}. These models capture contextualized representations of words and sentences by leveraging large-scale unsupervised training on vast amounts of text data. Instead of relying on explicit feature extraction methods, we utilize PLMs as feature extractors. These models encode the input text into dense vector representations, capturing intricate semantic and syntactic information. The resulting representations are then fed into an SVM classifier, which learns to discriminate between different classes. This approach has shown remarkable performance improvements in various NLP applications and has been validated in previous research~\cite{alhuzali2021predicting,peters2018deep}.

Building upon the concept of using PLMs, we consider the approach of fine-tuning these models. Fine-tuning involves training the PLMs on a task-specific dataset, allowing them to adapt to the specific classification objective. By fine-tuning PLMs, our aim is to capture task-specific nuances and optimize the models for improved classification performance. This approach has demonstrated promising results, as supported by research on universal language model fine-tuning~\cite{alhuzali2021spanemo,devlin2019bert,howard2018universal,Felbo}. Fine-tuning enables the models to effectively leverage their contextualized representations for accurate predictions.

In the realm of Artificial Intelligence (AI), the role of prompting engineering is pivotal in steering AI models towards producing outputs that align with user intentions. By providing specific input or instructions, the AI model can generate responses that are relevant, coherent, and in line with desired outcomes. This research paper focuses on harnessing the capabilities of GPT-3.5 and GPT-4. We closely examine the prompt used to explore the potential of these GPT models. The prompt provides instructions for the model to classify the text based on a predefined set of categories. To illustrate this process, here's an example:
\begin{itemize}
    \item Task[task prompt]: ``Your task is to analyze the question in the TEXT below from a patient and classify it with one or multiple of the seven labels in the following list:[``Diagnosis'', ``Treatment'', ...].

    \item Text[input text]: Be persistent with your doctor, and inform them of your concerns. Embrace the spiritual aspects and try to nurture them, as they will assist you in   transcendental thinking and relaxation. 

    \item Response[output]: {Guidance,   Emotional support}.
\end{itemize}

% We have explored three approaches for multi-label text classification: traditional feature extraction methods with SVM, utilizing pre-trained language models as feature extractors with SVM, and fine-tuning pre-trained large models. Each approach has its own merits and has been supported by prior research. Traditional feature extraction methods offer simplicity and interpretability, while pre-trained language models provide rich contextual information. Fine-tuning large models further enhances their performance by adapting them to the specific classification task. By comparing and analyzing these approaches, we contribute to a comprehensive understanding of text classification techniques, enabling researchers and practitioners to make informed decisions based on their requirements and constraints.

\subsection{Training Objective}
In the context of multi-label classification, the task involves selecting one or multiple labels. To determine if $P_i$ represents the correct question/answer type, we utilize the Sigmoid activation function. We utilize the binary cross entropy loss (BCE) when calculating the classification loss. The formalization of this loss function is as follows:

\begin{eqnarray}
Los{s_{BCE}} = \frac{1}{{\left| L \right|}}\sum\limits_{i = 1}^{\left| L \right|} {({Y_{ti}}\log ({P_{li}}) + (1 - {Y_{ti}})\log (1 - {P_{li}}))} 
\end{eqnarray}
where ${\left| L \right|}$ denotes the number of either question or answer types, depending on the specific task at hand. $Y_t$ corresponds to the ground truth label, while $P_{li}$ refers to the probability associated with the $i$-th label.

\subsection{Implementation Details}

For our experiments, we employed PyTorch~\cite{paszke2017automatic} as our framework of choice. The experiments were performed using a T4-GPU equipped with 15 GB of memory. For both extracting features from PLMs and fine-tuning them, we made use of the open-source Hugging-Face implementation~\cite{Wolf2019HuggingFacesTS}. We selected only three Arabic PLMs due to resource constrains: i.e., ``AraBERT'' developed by~\cite{antoun2020arabert}, ``MARBERT'' developed by~\cite{abdul-mageed-etal-2021-arbert}, and ``CAMeLBERT-DA'' developed by~\cite{inoue2021interplay}. According to the study conducted by~\cite{guo2024large}, both "AraBERT" and "MARBERT" demonstrate strong performance in detecting depression. Therefore, they are suitable candidates to be used in our benchmark study. For prompt engineering results, we used OpenAI API to conduct our experiments by considering two variants, i.e., GPT-3.5 and GPT-4.

For evaluation, given that the two tasks involve multi-label classification, we employed three widely used metrics~\cite{alhuzali2021spanemo, mohammad2018semeval}: i.e., Micro F1-score, weighted F1-score, and Jaccard index score. The presence of a class imbalance issue in the MentalQA dataset is worth mentioning. To ensure a more equitable evaluation of the model's performance, we opt for the weighted-F1 score instead of macro-F1, as it takes into account the varying class distribution and mitigates the effects of imbalanced classes.

To ensure consistency, all models employed in this paper were trained using identical hyper-parameters and a fixed initialization seed. The hyper-parameter settings consisted of a feature dimension of 786, a batch size of 8, a dropout rate of 0.1, an early stop patience of 10, and a training duration of 15 epochs. For optimization, we selected the Adam optimizer~\cite{kingma2014adam} with a learning rate of $2e\text{-}5$. Furthermore, we leveraged scikit-learn~\cite{scikit-learn} as a valuable resource for implementing the SVM algorithm as well as extracting both counting-based and TF-IDF features. It is important to note that all aforementioned models were tuned exclusively on the validation set. Table~\ref{hyper} provides a summary of the aforementioned hyper-parameters.

\begin{table}[h]
\centering 
\caption{A summary of hyper-parameters utilized in this work.}\label{hyper}
\begin{tabular}{lc}\toprule
\textbf{Parameter} & \textbf{Value}\\ \midrule
{dimension}  & $768$\\
Batch-size & $8$  \\
Dropout   & $0.1$\\
Early-stop-patience & $10$ \\
\#epochs & $15$ \\
learning rate & $2e\text{-}5$ \\
Optimizer & Adam \\
\bottomrule
\end{tabular}
\end{table}

\subsection{Experimental Results}\label{results}

% Dicuss first types of learning representation 

% Discuss setup of training and testing

Table~\ref{res_}\footnote{This study focused on exploring the MentalQA dataset and, it conducted the first set of experiments on this particular dataset. As a result, there are currently no existing SOTA models available for comparison.} shows the performance of the three explored approaches on the tasks of classifying question and answer types. The evaluation metrics used to assess the performance include micro F1-score, weighted F1-score, and Jaccard score. We also compared the results against the following baselines, i.e., random and most frequent class. The random baseline approach serves as a simple benchmark by assigning question and answer types randomly, resulting in low performance across all evaluation metrics. This highlights the need for more sophisticated models in question and answer classification. The most frequent class baseline approach selects the most common question and answer types as predictions for all samples, outperforming the random baseline but still falling short compared to more advanced models.
% Additionally, we examined three distinct setups to accomplish our objective. Firstly, we trained and tested the model exclusively on questions. Secondly, we trained on questions and tested on answers. Lastly, we trained and tested the model solely on answers. The purpose of training on questions and testing on answers was to investigate the possibility of utilizing patient inquiries to identify potential responses with respect to the selected categories. This information would assist PLMs in determining the most appropriate response to patients' queries. 

% \usepackage{multirow}
\begin{table}[htp]
\centering
\caption{The results of multi-label Q/A types classification on the test set of MentalQA. Best: \textbf{bold}.}
\label{res_}
\resizebox{\textwidth}{!}{%
\begin{tabular}{lcccccc}
\toprule
\multirow{3}{*}{\textbf{Model}} & \multicolumn{3}{c}{\textbf{Results of Q Classification}} & \multicolumn{3}{c}{\textbf{Results of A Classification}} \\ 
% &
  % \multicolumn{3}{c|}{\textbf{Trained   and Test on Q}}  &
  % \multicolumn{3}{c}{\textbf{Trained   and Test on A}} \\
 & F1-Micro & F1-Weighted & Jaccard-Score & F1-Micro &  F1-Weighted &  Jaccard-Score 
  \\\midrule
\multicolumn{7}{c}{\textbf{Baseline}}                                                \\ \midrule
Random               & 0.33  & 0.26 & 0.21  & 0.50 & 0.45 & 0.34 \\
Most   common class  & 0.62 & 0.75 & 0.49 & 0.79 & 0.83 & 0.67 \\ \midrule
\multicolumn{7}{c}{\textbf{SVM}}                                                     \\ \midrule
SVM   (Occurrences)  & 0.83  & 0.84 & 0.78  & 0.91 & 0.92 & 0.87 \\
SVM   (Count)        & 0.83  & 0.84 & 0.78  & 0.93 & 0.93 & 0.90 \\
SVM   (TF-IDF)       & 0.84  & 0.85 & 0.79  & 0.93 & 0.93 & 0.90 \\
 \midrule
\multicolumn{7}{c}{\textbf{PLMs As Feature   Extractor}}                             \\ \midrule
SVM   (AraBERT)      & 0.80  & 0.81 & 0.74  & 0.92 & 0.92 & 0.89 \\
SVM   (CAMelBERT-DA) & 0.81  & 0.81 & 0.79  &   0.92  &    0.93  &  0.91    \\ %0.929	0.931	0.91
SVM   (MARBERT)      & 0.81  & 0.81 & 0.79  & 0.92 & 0.92 & 0.89 \\ \midrule
\multicolumn{7}{c}{\textbf{Fine-tuning PLMs}}                                        \\ \midrule
AraBERT              & 0.84  & 0.82 & 0.79  & 0.94 & 0.94 & 0.92 \\
CAMelBERT-DA         & 0.85  & 0.83 & 0.80  &  0.94    &  0.93    &  0.92    \\ %0.94	0.938	0.92
MARBERT              & \textbf{0.85}  & \textbf{0.85} & \textbf{0.80}  & \textbf{0.95} & \textbf{0.95} & \textbf{0.94} \\  \midrule
\multicolumn{7}{c}{\textbf{Prompt Engineering}}                                        \\ \midrule
GPT-3.5 (Zero-Shot)   & 0.59  & 0.55 &  0.45 & 0.42 & 0.40 & 0.39 \\ %0.585 0.549 0.45 %%% 0.419 0.403 0.386
GPT-4  (Zero-Shot)   & 0.57  & 0.54 &  0.44  & 0.43 & 0.41 & 0.40 \\ % 0.566 0.535 0.439 %%% 0.419 0.403 0.386
% GPT-4o  (Zero-Shot)   & 0.56   & 0.53  & 0.44 &  &  &  \\ % 0.655 0.597 0.52 %%%0.663 0.676 0.554
GPT-3.5 (Few-Shot-3)   &  0.66 & 0.61 &  0.53  & 0.66 & 0.68 & 0.57 \\ %0.664 0.611 0.526 %% %0.663 0.679 0.564
GPT-4  (Few-Shot-3)   &  0.66 & 0.60 & 0.52   & 0.66 & 0.68 & 0.56 \\ % 0.655 0.597 0.52 %%%0.663 0.676 0.554
% GPT-4o  (Few-Shot-3)   &   &  &  &  &  &  \\ % 0.655 0.597 0.52 %%%0.663 0.676 0.554

\bottomrule
\end{tabular}
}
\end{table}

The first explored approach involves traditional feature extractor methods utilizing SVM with different feature representations. These models leverage occurrence-based, count-based and TF-IDF representations to capture word presence/absence, word frequency and importance in the input data, respectively. The SVM classifier utilizes these features to achieve higher performance scores compared to the baselines. The TF-IDF representation generally outperforms the occurrence-based and count-based representation, indicating that weighting words based on importance enhances model performance. Furthermore, it is worth noting that the SVM with TF-IDF features exhibited competitive performance to PLMs, as supported by previous research~\cite{wahba2022comparison}. This outcome is to be expected, considering the limited size of the training set. Nevertheless, it is important to acknowledge that this situation might alter if the dataset expands to a larger scale.

% For example, the TF-IDF with SVM achieved an F1-Micro score of 0.84, an F1-Macro score of 0.85, and a Jaccard Score of 0.79 for question classification. In answer classification, it obtained an F1-Micro score of 0.89, an F1-Macro score of 0.90, and a Jaccard Score of 0.85. 

The second approach utilizes PLMs as feature extractors in combination with the SVM classifier. AraBERT, CAMelBERT, and MARBERT, three Arabic PLMs, are employed to extract contextualized representations of the input data. These models achieve competitive performance scores for both question and answer types, benefiting from their ability to capture the semantic meaning of the text. MARBERT outperformed both Arabic PLMs in almost all setups and metrics.  

The third approach Fine-tuned PLMs, which takes it a step further by adapting the PLMs specifically to the tasks of question and answer classification. Through fine-tuning the three Arabic PLMs, these models achieve even higher performance scores compared to using the PLMs solely as feature extractors. Fine-tuning allows the models to learn task-specific patterns and optimize their performance on the given dataset. The results of this approach demonstrates that Fine-tuning MARBERT achieved the best performance.

The fourth approach explores the use of prompting PLMs such as GPT-3.5 and GPT-4. The objective was to evaluate their performance in two different scenarios: zero-shot learning and few-shot learning, specifically focusing on question and answer classification. The results indicate that the few-shot learning setting, utilizing only three instances of labeled data, yields better outcomes compared to the zero-shot learning setting. Across all three metrics, there was an overall performance improvement of up to 7\%. This finding suggests that few-shot learning holds great potential for future research aimed at enhancing model performance while working with limited labeled data, which is often the case in real-world applications. By leveraging this approach, it becomes possible to achieve notable advancements even with a small amount of available data.

% for question classification, achieving an F1-Micro score of 0.81, an F1-Macro score of 0.81, and a Jaccard Score of 0.79. However, AraBERT achieved better performance for answer classification, with an F1-Micro score of 0.89, an F1-Macro score of 0.89, and a Jaccard Score of 0.84.

% , with an F1-Micro score of 0.85, an F1-Macro score of 0.85, and a Jaccard Score of 0.80 for question classification. In answer classification, it obtained an F1-Micro score of 0.90, an F1-Macro score of 0.90, and a Jaccard Score of 0.86.

The performance of the four discussed approaches for classifying question and answer types was evaluated and compared to simple baselines. The first approach utilized traditional feature extractors with SVM, achieving higher performance than the baselines by capturing word frequency and importance through count-based and TF-IDF representations. The second approach employed PLMs as feature extractors, with AraBERT, CAMelBERT, and MARBERT achieving competitive scores by capturing semantic meaning. Fine-tuning the PLMs further improved performance by adapting them specifically to the question and answer classification tasks, resulting in even higher scores. The fourth approach involves promoting GPT models in two different settings, where the second one exhibited a strong performance. The findings from these results demonstrate the usefulness of each model employed and the utilization of contextualized representations for achieving accurate classification of question and answer types. These results also emphasize the crucial role played by the selected models and the value of incorporating contextual information in effectively categorizing question and answer types. The results obtained from GPT indicate that few-shot learning has the potential to substantially enhance model performance, even when working with a limited number of labeled data instances.

% In the second setup, the model was initially trained on questions and subsequently tested on answers. The obtained results in this case were not as high as when the model was trained on answers directly. However, despite the lower performance, the model still demonstrates reasonable capability. This indicates that questions contain valuable cues that can prompt the appropriate responses from doctors when addressing patients' inquiries. Moreover, this setup can also be utilized to condition the model's response generation by taking into account the relevant category. For instance, incorporating the appropriate category information can further enhance the accuracy and relevance of the generated responses for patients' inquires.

\section{Evaluation of The Results}\label{analysis}
% What we can add is the followings:
% 1. Modify training objective to handle both unbalanced or multi-label cases.
% 2. Perforum Semi-Supervised learning
% 3. instruction tuning or prompting engineering

In this section, we conducted a series of comprehensive analyses that contribute to a deeper understanding of our experiments undertaken in this paper. These analyses encompassed various aspects and aimed to shed light on key factors influencing model performance. In the first analysis, our focus centered on evaluating the effect of fine-tuning PLMs compared to not fine-tuning them. By exploring both scenarios, we gained crucial insights into the benefits and trade-offs associated with fine-tuning PLMs. This analysis enabled us to make informed decisions regarding the optimal approach for our specific task. The second analysis shifted towards investigating the influence of data size on the performance. We assessed the influence of data availability on the performance of the models by systematically adjusting the size of the training set. This examination provided valuable insights into the relationship between data quantity and model performance. Furthermore, we conducted a detailed case study as part of the third analysis. The goal of this investigation was to identify potential errors or limitations in the employed models. By examining their performance and scrutinizing any discrepancies or inaccuracies, we aimed to identify areas for improvement and guide future work in addressing these issues effectively.

\subsection{Effect of fine-tuning PLMs}
\begin{figure}[h]
\centering
{\includegraphics[width=0.9\linewidth]{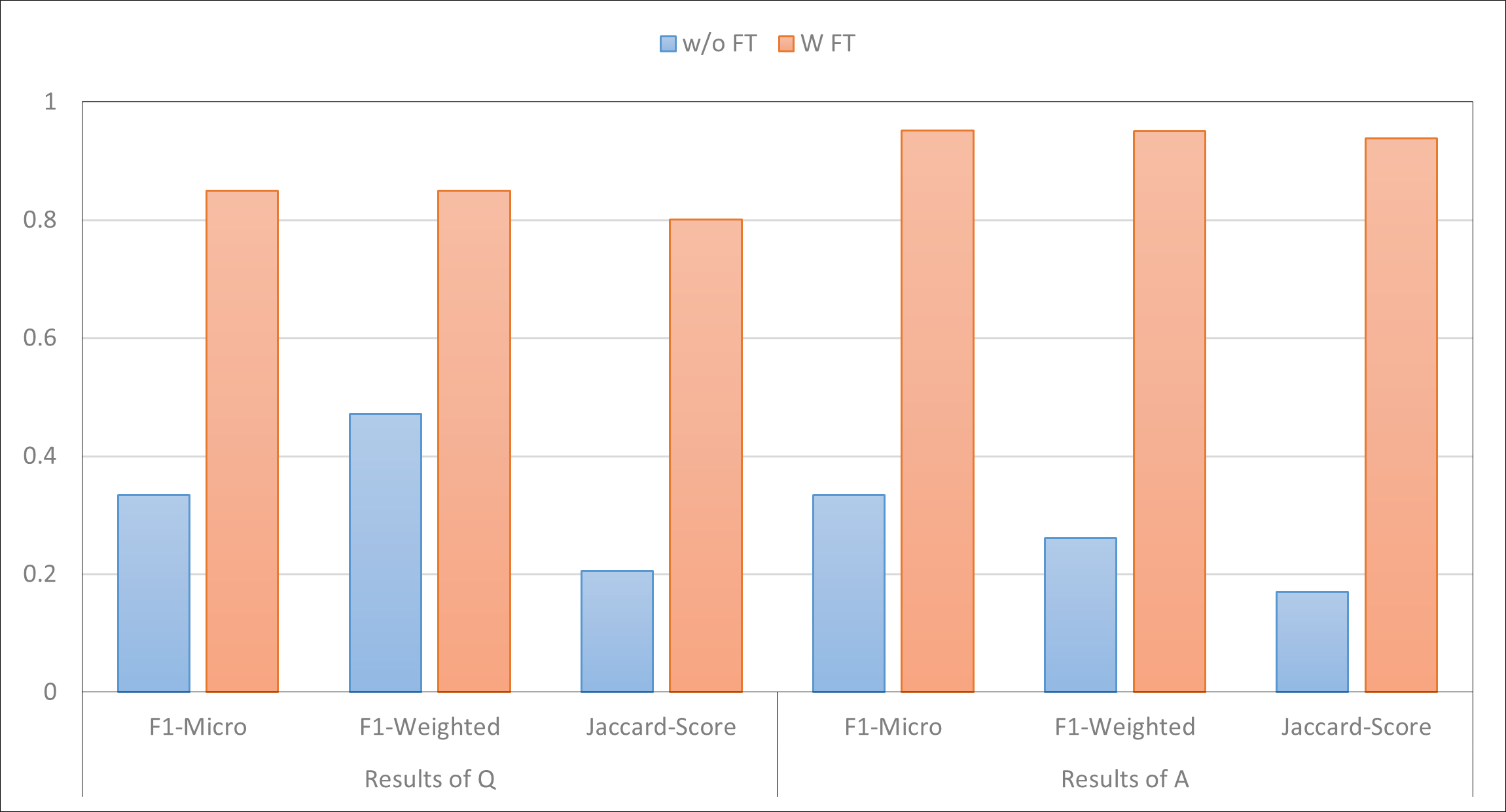}} \\
\caption{Illustrating the impact of fine-tuning PLMs compared to not fine-tuning them. The x-axis of the plot represents the metrics employed in the paper, while the y-axis represents the corresponding model scores. Additionally, the color bars within the plot indicate with fine-tuning (w/ FT) vs without fine-tuning (w/o FT).}\label{effect2}
\end{figure}

In this section, we conducted an analysis to assess the effect of fine-tuning PLMs compared to not fine-tuning them. The results presented in Figure~\ref{effect2} focus on the performance of MARBERT. It was observed that this models achieved very low performance when it was not fine-tuned using the train set of the MentalQA dataset. This finding is consistent with prior research, which indicates that PLMs are typically trained on general domain data~\cite{gururangan2020don}. Consequently, fine-tuning becomes an essential step in order to achieve good performance, even with a small amount of labeled data. However, fine-tuning MARBERT exhibited the best performance. These results clearly indicate that fine-tuning the models significantly improves its performance in both question and answering types classification tasks, specifically, when trained on task-specific data~\cite{devlin2019bert,alhuzali-ananiadou-2019-improving,howard2018universal}. We now turn to discussing the effect of fine-tuning the same model using varying data sizes.

% Since both AraBERT and CAMelBERT obtain identical trends, they were remove The same trends were obseved for AraBERT and CAMelBERT; they were remove for 

\subsection{Effect of Zero-Shot vs Few-Shot learning}
\begin{figure}[h]
\hfill
\begin{center}
\subfigure[The results of question types]{\includegraphics[width=8cm]{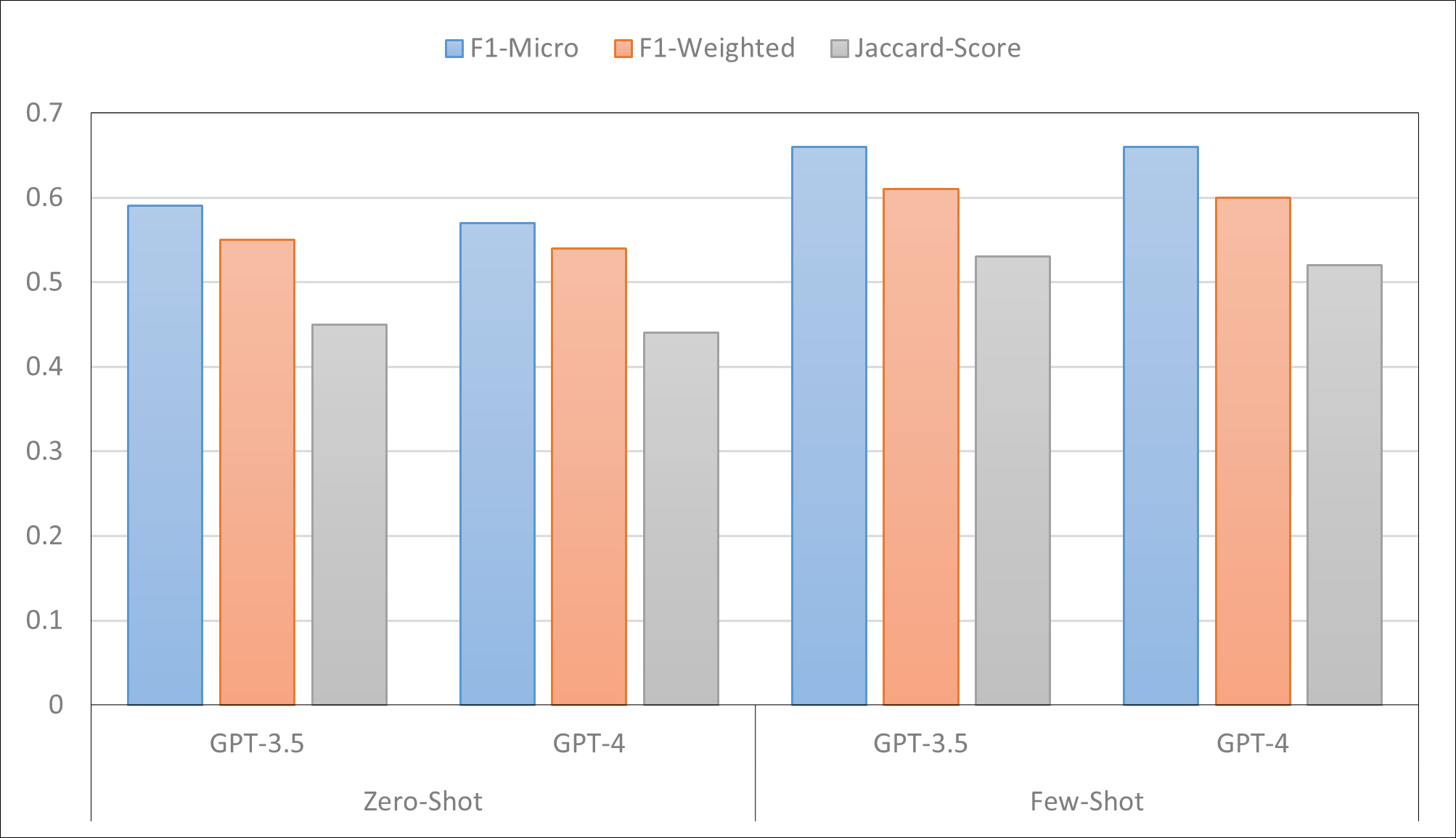}}
% \hfill
\subfigure[The results of answer types]{\includegraphics[width=8cm]{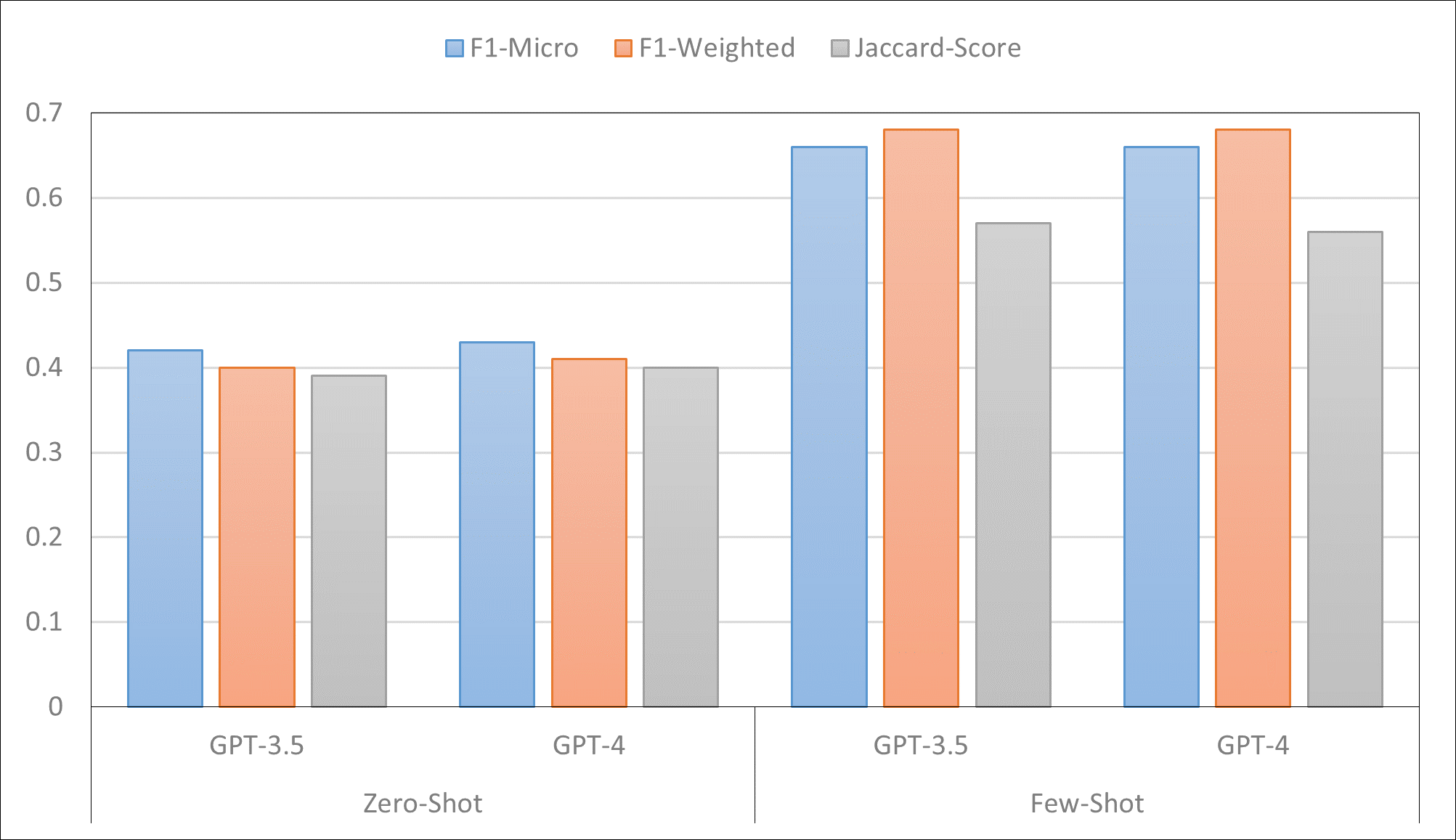}}
\hfill
\caption{Both plots depict the impact of few-shot learning on model performance.} \label{gpt_results}
\end{center}
\end{figure}

In this analysis, our objective was to assess the effectiveness of prompting PLMs like GPT as shown in Figure~\ref{gpt_results}. To achieve this, we chose two prominent large language models, namely GPT-3.5 and GPT-4, and evaluated their performance in two specific learning settings: zero-shot learning and few-shot learning, utilizing just three labeled examples. Surprisingly, the results revealed that GPT-3.5 outperformed GPT-4 in both scenarios. Notably, even with the limited labeled data, GPT-3.5 showcased superior performance, showcasing its ability to leverage small labeled datasets effectively. These findings suggest that GPT-3.5 exhibits a strong capacity for understanding and generalizing from limited instances, surpassing the advancements made in GPT-4. This implies that when dealing with scenarios involving small textual labeled datasets like the ones used in this study, GPT-3.5's performance in few-shot learning surpasses the capabilities of GPT-4. Understanding the strengths and limitations of different language models is crucial in determining the most suitable approach for specific natural language processing tasks.

In the zero-shot setting, GPT-3.5 demonstrated a micro F1-score of 0.59, a weighted F1-score of 0.55, and a Jaccard score of 0.45 for question classification, indicating its ability to generate reasonably accurate predictions without any specific training on the given task. However, when transitioning to the few-shot setting with only three labeled examples, GPT-3.5 experienced a noticeable increase in performance. It achieved a micro F1-score of 0.66, a weighted F1-score of 0.61, and a Jaccard score of 0.53. This improvement suggests that even with a minimal amount of labeled data, GPT-3.5 was able to leverage the provided examples and extract meaningful patterns, resulting in higher precision and recall rates. The significant boost in performance highlights the model's capacity to adapt and utilize limited supervised information effectively, showcasing its ability to generalize and make more accurate predictions in the few-shot learning scenario. The benefits of few-shot learning in boosting performance were also observed in the context of answer classification.

\subsection{Effect of data size on performance}

\begin{figure}[h]
\centering
\subfigure[The results of question types]{\includegraphics[width=8cm]{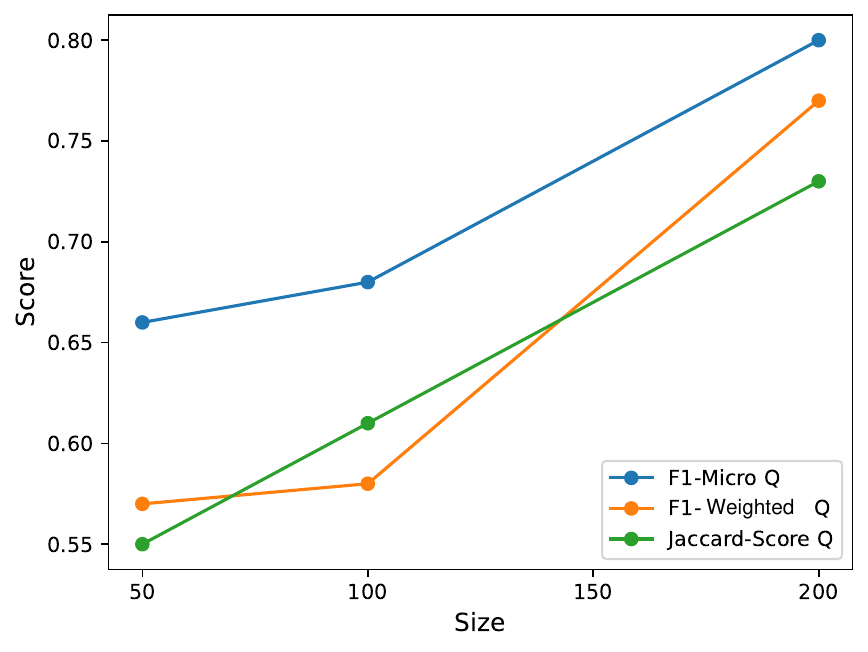}}
% \hfill
\subfigure[The results of answer types]{\includegraphics[width=8cm]{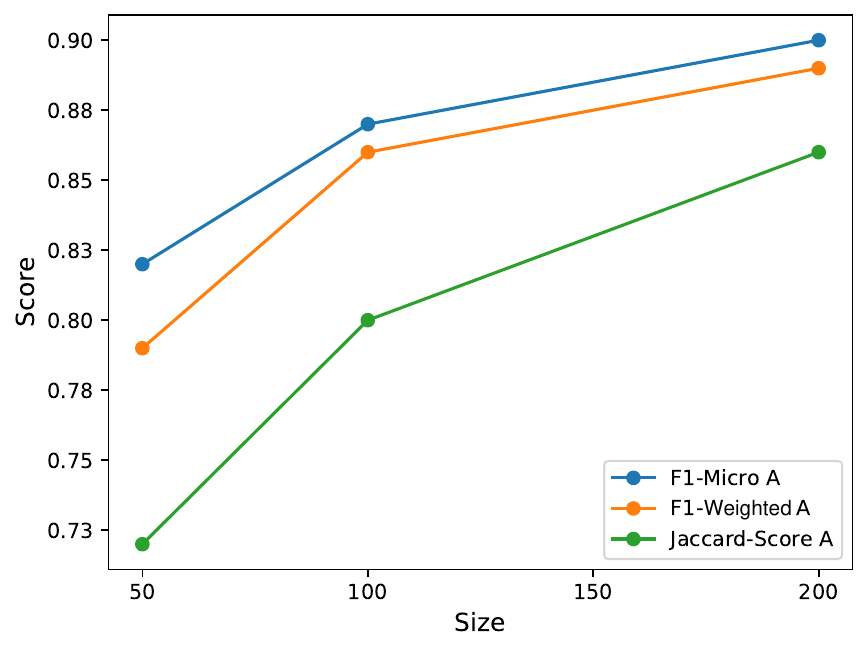}}
\hfill
\caption{Both plots depict the impact of data size on model performance. The x-axis in the plot indicates the number of samples utilized for training, while the y-axis corresponds to the score of each metric.} \label{effect}
% \end{center}
\end{figure}

The analysis conducted in this section involves evaluating the performance of the MARBERT model\footnote{We chose this model because it achieved almost the best performance on all metrics.} for question types and answer types classification. The study examines the impact of varying the size of the training data on the model's performance. The evaluation utilizes three metrics, i.e., F1-Micro, F1-Macro, and Jaccard-Score - to assess the model's performance. The analysis is presented in Figure~\ref{effect} through two plots, where the top plot represents question types classification and the bottom plot represents answer types classification.

Figure~\ref{effect} displays the influence of data size on model performance. The x-axis in both plots represents the size of the data, ranging from 50 samples to 200 samples, while the y-axis represents the corresponding performance scores. The analysis reveals a clear relationship between data size and model performance. The results obtained from the plots indicate that increasing the data size positively affects the model's performance in both question and answer types classification. Notably, as the data size increases, there is a substantial improvement in performance. For question types classification, the F1-Micro score rises from 0.65 to 0.80, the F1-Macro score increases from 0.57 to 0.77, and the Jaccard score improves from 0.55 to 0.73. Similarly, the results of answer types classification display a similar trend, confirming the positive impact of larger datasets on model performance.

The analysis results emphasize the importance of data size in attaining improved performance scores for both question and answer types classification tasks. These findings demonstrate the effectiveness of harnessing data size as a means to enhance the model performance. By increasing the amount of available data, the model's performance can be substantially enhanced, leading to more accurate results in both question and answer types classification tasks. 

% In additiona, the use of a PLM adopted the task of Q/A types classification indicates the positive effect of 

% outcomes of a PLM

% \begin{figure}[!t]
% \centering
% % \subcaptionbox{The results of Question types\label{Qr}}
% {\includegraphics[width=0.5\linewidth]{figs/results_q.pdf}} \\
% % \subcaptionbox{The results of Answer types\label{Ar}}
% {\includegraphics[width=0.5\linewidth]{}}
% \caption{Both plots depict the impact of data size on model performance. The x-axis in the plot indicates the number of samples utilized for training, while the y-axis corresponds to the score of each metric.}\label{effect}
% \end{figure}

\subsection{Case Study}

Upon analyzing the model predictions, we identified several noteworthy issues as shown in Table~\ref{case-study}. Firstly, the model's inability to accurately predict the emotional support label in the first three instances highlights its limitations in capturing the intricate emotional nuances within the text. Extensive research has emphasized the significance of emotions in comprehending and addressing mental health concerns~\cite{zhang2023emotion, alhuzali2022neural}. The model's errors in recognizing emotions indicate the need for further improvement in this aspect, as accurate emotional understanding is crucial for providing appropriate guidance.

% \usepackage{multirow}
% \usepackage[normalem]{ulem}
\begin{table}[h]
\centering
\caption{Comparing model's predictions to the actual labels, specifically in the context of answer and question classification. The terms ``info'', ``guid'', ``emo'', ``diag'', ``treat'', ``prov'', and ``H.life'' are abbreviations that stand for information, guidance, emotional support, diagnosis, treatment, health provide choice, and health lifestyle, respectively. The translated text is the result of utilizing ChatGPT to translate the original text from Arabic to English. We excluded the original text for the sake of sapce.}
\scalebox{0.8}{
\label{case-study}
\begin{tabular}{lp{40em}ll}
\toprule
% \multicolumn{2}{c}{} & \multicolumn{2}{c}{Labels} \\
\# & {Translated Text} & {Actual} & {Prediction} \\ \midrule
\multicolumn{4}{c}{Examples representing Answers}\\ \midrule 
1 & Be persistent with your doctor, and inform them of your concerns. Embrace the   spiritual aspects and try to nurture them, as they will assist you in   transcendental thinking and relaxation. & {Guid,   Emo} & {Guid} \\ %\midrule
2 & Good   evening. It is necessary to consult a mental health professional to assess   your condition. In the meantime, think positively about yourself and   reconcile with yourself, loving yourself. & {Guid,   Emo} & {Guid} \\
3 & The sadness will end, and you will live in complete happiness, God willing. I advise you to speak to a specialist in psychological therapy, as there are modern methods and wonderful medications that can help you, God willing, to live happily. Many people go through tough times and overcome them, and you are one of them. Do not suppress your sadness, and consult someone you trust. & Guid, Emo & Guid \\ 
\midrule
% 4 & Yes, there is   treatment, I advise you to visit a psychiatrist to help you. & Guid & Info, Guid \\ %\midrule
\multicolumn{4}{c}{Examples representing Questions}\\ \midrule 
4 & I went to a psychiatrist two years ago and received treatment there. The symptoms disappeared, and I felt much better, so I stopped the treatment. However, now the sleep disturbances and anxiety have returned. What should I do? Should I go back to the doctor? & diag, treat, prov &  treat \\ 

5 & ``I am a 40-year-old man who has never been married before, but now I am considering getting married. However, some people advise me against getting married at this age, saying that I won't be able to live a happy life after reaching this age''. & H.life & diag, treat \\

6 & The withdrawal symptoms of escitalopram are intense for me when I stopped taking it due to fear of its impact on pregnancy. However, the symptoms were strong, and I visited the doctor feeling unwell. My blood tests and urine tests came out normal, but I experienced a setback after discontinuing the medication two months ago. I am unable to visit the psychiatrist, and I have been using a dose of 10mg and then 20mg for approximately two and a half years. I need urgent help. & diag, treat & treat \\
\bottomrule
\end{tabular}}
\end{table}

Furthermore, the last three example sheds light on a common challenge in multi-label classification, which involves defining clear boundaries between certain labels. As mentioned in the works of~\cite{zhang2023phq} and~\cite{alhuzali2021spanemo}, the relationships between labels directly impact the performance of the model. Addressing this challenge and considering label relationships could potentially enhance the model's performance. In numerous cases, the labels ``diagnosis and treatment'' are frequently encountered. Exploring the dynamics between labels and developing strategies to handle them effectively in multi-label question classification could serve as an interesting area for future research and improvement.

\section{Discussion}\label{disc}

Our experiments yielded promising results, highlighting the potential of fine-tuned Arabic PLMs for Arabic mental health support systems. Fine-tuned models, like MARBERT, consistently outperformed traditional feature extraction methods with SVM for both question and answer classification tasks and that is evident in the prior work \cite{almazrouei-etal-2023-alghafa, abdelali-etal-2024-larabench}. This margin of improvement suggests that PLMs are adept at capturing the nuances and context of Arabic language specific to the domain of mental health. Notably, fine-tuning the PLMs on a dataset curated specifically for mental health questions and answers resulted in a significant performance boost compared to the non-fine-tuned model. This finding underscores the importance of domain adaptation – tailoring the PLMs to the specific domain through targeted training on relevant data. Additionally, our exploration revealed a clear correlation between dataset size and model performance, echoing prior research findings in NLP that larger datasets consistently yield better results \cite{althnian2021impact}. This highlights the crucial role of data availability in this field, where the under-representation of Arabic mental health resources necessitates further data collection efforts.

In addition, our study also sheds light on an interesting dynamic concerning general domain Arabic PLMs (AraBERT, CAMelBERT, MARBERT) when applied to specific tasks. While these models achieved promising performance on the MentalQA dataset, they might benefit from further training on even more relevant mental health corpora. This highlights the potential for developing Arabic PLMs specifically trained on mental health data, like a  "MentalBERT" or "MentalRoBERTa" model ~\cite{ji2022mentalbert}.  Such models could be pre-trained on a massive dataset of Arabic text specifically tailored to the mental health domain, including mental health questions, clinical notes, support group discussions, and mental health resources. This targeted training could allow the PLMs to develop a deeper understanding of the nuances and terminology used in mental health conversations, potentially leading to even better performance on downstream classification tasks within this domain.

The results from the fine-tuned models on the mental health dataset lays the groundwork for exploring the effectiveness of such domain-specific Arabic PLMs for mental health applications. This aligns with calls for further research on the application of NLP for mental health tasks, particularly in under-resourced languages like Arabic ~\cite{zhang2022natural}. Future research can investigate how these models can be further optimized to achieve specific goals, such as improved accuracy in type of questions sought or enhanced ability to identify users at risk. Additionally, exploring techniques like semi-supervised learning or leveraging unlabeled data through methods like contrastive learning could potentially enhance model performance while mitigating the limitations of dataset size. By pursuing these research directions, we can contribute to the development of more robust and effective Arabic language models for mental health support systems, ultimately improving accessibility to mental health resources for Arabic-speaking communities.

\subsection{Implications} 
Our study using the MentalQA dataset demonstrates the potential of PLMs for Arabic mental health support systems, particularly in question and answer classification tasks. These findings hold significant implications for the future of mental health resources in general and more specifically in Arabic language. By effectively classifying questions and answers, PLMs can facilitate intervention assistance by directing users towards appropriate resources or connecting them with potential therapists based on their specific needs. This targeted approach can significantly reduce the time and effort required for individuals seeking help.

PLMs can offer non-clinical guidance and support around the clock, even in the absence of readily available human therapists. This can include providing access to educational materials on mental health topics, self-help strategies, or coping mechanisms for managing symptoms.  Early intervention through readily available PLM-powered support systems can play a crucial role in promoting mental well-being.

Future advancements in PLMs could enable personalized interactions within mental health support systems. By analyzing user history and preferences, PLMs could tailor information and support to individual needs. This could lead to more effective interventions and improved mental health outcomes.

%By effectively classifying questions and answers, PLMs can facilitate intervention assistance by directing users towards appropriate resources or connecting them with human therapists. PLMs can also provide non-clinical guidance and support, such as educational materials or coping mechanisms, for individuals experiencing mental health challenges. Future advancements in PLMs could enable personalized interactions, tailoring support to individual needs and preferences. 

\subsection{Limitations and Ethical Consideration} 
In light of our experimental findings on the MentalQA dataset, we have obtained promising results in discerning question and answer types. However, it is important to acknowledge that our study encompasses certain potential limitations and highlights avenues for future exploration. Our experiments were not designed for diagnostic purposes, but rather aimed at offering an estimation for categorizing Q/A types, thereby facilitating intervention assistance and providing guidance for non-clinical applications.

\section{Conclusion}\label{conc}
In conclusion, this study explored the effectiveness of classical machine learning models and pre-trained language models (PLMs) on the MentalQA dataset, the Arabic mental health question-answering dataset featuring question-answering interactions. While traditional feature extractors with SVM achieved a strong performance, PLMs like MARBERT and AraBERT demonstrated even better performance due to their ability to capture semantic meaning. These results suggest that PLMs hold significant promise for Arabic mental health question-answering, providing intervention services or connecting patients with resources. Furthermore, the experiment of prompt engineering has provided valuable insights into the advantages of few-shot learning when compared to zero-shot learning. Through our exploration, we were able to enhance question classification results by 12\% and improve answer classification results by 45\%. These findings demonstrated the potential of PLMs in the domain of Q\&A classification for mental health care. By leveraging PLMs, we can facilitate the development of accessible and culturally sensitive resources specifically tailored to Arabic-speaking populations. This research contributes to the understanding of the feasibility of utilizing PLMs for Arabic mental health care, ultimately aiming to address the needs of this demographic.

\section*{Summary Table}

\textbf{What was already known on the topic:}
\begin{itemize}
    \item There's a growing trend of developing domain-specific models, particularly in medical domain.
    \item Research on utilizing pre-trained language models for mental health, especially in underrepresented languages, is scarce.
\end{itemize}

\textbf{What this study added to our knowledge:}

\begin{itemize}
    \item This study demonstrates the potential of pre-trained language models for classification of questions and answers in mental health for Arabic language.
    \item The research highlights the effectiveness of prompt engineering, showing the benefits of few-shot learning compared to zero-shot learning for this task.
\end{itemize}

\printcredits

% \textbf{H.A.}: Methodology, Software, Validation, Formal analysis, Visualization, Writing - original draft.
% \textbf{A.A.}: Methodology and Writing - review \& editing.

\section*{Declaration of competing interest}
The authors declare that they have no competing interests.

\section*{Acknowledgment}
We extend our appreciation to Google Colab for enabling us to conduct the experiments presented in this paper. \textbf{Declaration of generative AI and AI-assisted technologies in the writing process:} During the preparation of this work, we used ChatGPT, an AI chatbot developed by OpenAI, in order to improve our written work. After using this tool/service, we reviewed and edited the content as needed and take full responsibility for the content of the publication.

\bibliographystyle{unsrt}
% \bibliographystyle{srt}
% \bibliographystyle{apacite}
\bibliography{cas-refs}

\end{document}